\documentclass[preprint,12pt]{elsarticle}

\usepackage{amssymb}
\usepackage{amsmath}

 \usepackage{lineno}

\usepackage{hyperref} 
\usepackage{algorithm}%
\usepackage{algorithmicx}%
\usepackage{algpseudocode}%
\usepackage{listings}%
\usepackage[T1]{fontenc}

\begin{document}

\begin{frontmatter}



\title{Towards consistency of rule-based explainer and black box model - fusion of rule induction and XAI-based feature importance}



\author[polsl]{Micha\l{} Kozielski}
	
\author[polsl]{Marek Sikora\corref{mycorrespondingauthor}}
	\ead{marek.sikora@polsl.pl}

\author[emag]{\L{}ukasz Wawrowski}

\cortext[mycorrespondingauthor]{Corresponding author}

\affiliation[polsl]{organization={Department of Computer Networks and Systems, Silesian University of Technology},
            addressline={Akademicka 16}, 
            city={Gliwice},
            postcode={44-100}, 
            country={Poland}}

\affiliation[emag]{organization={\L{}ukasiewicz Research Network – Institute of Innovative Technologies EMAG},
            addressline={Leopolda 31}, 
            city={Katowice},
            postcode={40-189}, 
            country={Poland}}

\begin{abstract}
Rule-based models offer a human-understandable representation, i.e. they are interpretable. For this reason, they are used to explain the decisions of non-interpretable complex models, referred to as black box models. The generation of such explanations involves the approximation of a black box model by a rule-based model.
To date, however, it has not been investigated whether the rule-based model makes decisions in the same way as the black box model it approximates. Decision making in the same way is understood in this work as the consistency of decisions and the consistency of the most important attributes used for decision making. 
This study proposes a novel approach ensuring that the rule-based surrogate model mimics the performance of the black box model. The proposed solution performs an explanation fusion involving rule generation and taking into account the feature importance determined by the selected XAI methods for the black box model being explained. The result of the method can be both global and local rule-based explanations.
The quality of the proposed solution was verified by extensive analysis on 30 tabular benchmark datasets representing classification problems. Evaluation included comparison with the reference method and an illustrative case study. In addition, the paper discusses the possible pathways for the application of the rule-based approach in XAI and how rule-based explanations, including the proposed method, meet the user perspective and requirements for both content and presentation. The software created and a detailed report containing the full experimental results are available on the GitHub repository (\href{https://github.com/ruleminer/FI-rules4XAI}{https://github.com/ruleminer/FI-rules4XAI}).
\end{abstract}

\begin{keyword}
XAI \sep feature importance \sep object-related rule induction \sep rule-based explanations
\end{keyword}

\end{frontmatter}

\section{Introduction}
\label{sec:introduction}

Developments in technology are enabling (and motivating) the creation of increasingly complex artificial intelligence (AI) and machine learning (ML) models. Therefore, many of the most recent methods generate models that are not interpretable. This means, that the user who is the recipient of the model, i.e. a human, is not able to understand on what basis the generated model makes a decision. Consequently, models with such a representation are referred to as black box. In contrast, models that are interpretable to a human are referred to as white box (glass box). The representation of a phenomenon that is created during their generation is interpretable for a human, i.e. the user of such a model is able to understand what the model's decision is based on.

The models generated by AI/ML methods are now widely used and their decisions affect an increasing number of users - not only data analysis specialists but experts in other fields and ordinary users. A need has therefore emerged to explain the decisions of black box models. This need is not only expressed by users of AI/ML systems but also by legislators \cite{Goodman_Flaxman_2017}. In response to this demand, a number of explainable artificial intelligence (XAI) methods have been developed \cite{BARREDOARRIETA202082,adadi2018peeking,molnar2022,biecek2021explanatory}.   
Thus, on the one hand, the need to explain black box model decisions is clearly stated and the involvement of the data science community in the development of XAI methods is clear. On the other hand, there are voices \cite{Rudin_2022} indicating problems concerning the quality of explanations (fidelity) and their interpretability. It is emphasised that the approximation of a complex (black box) model by an interpretable model is not prefect, as it is mostly associated with simplification. This may lead to limited trust in the explanation. A second problem is that the explanatory approximator does not have to mimic the operation of the black box model. Therefore, the resulting explanations may be based on different features and lead to incorrect conclusions.

The motivation for the present study is related to the above issues. To date, to the best of the authors' knowledge, no research has been undertaken on whether when black box models and their rule-based white box approximators make the same or similar decisions they do this on the same basis. There has been numerous studies on the use of so-called "surrogate models" \cite{10.1145/3531146.3534639}, which are approximators of complex models. Typically, the convergence of the performance of a back box model and its approximator is measured by the error made by both models on training or test datasets. Less commonly, the consistency of decisions made is tested (calculating e.g. Cohen's Kappa on the decisions of black box and white box models). Another possibility is the use of synthetic explainable classifier generators \cite{GUIDOTTI2021103428} which allows the comparison of local XAI methods.

\begin{figure}[!htb]
    \centerline{\includegraphics[width=0.95\textwidth]{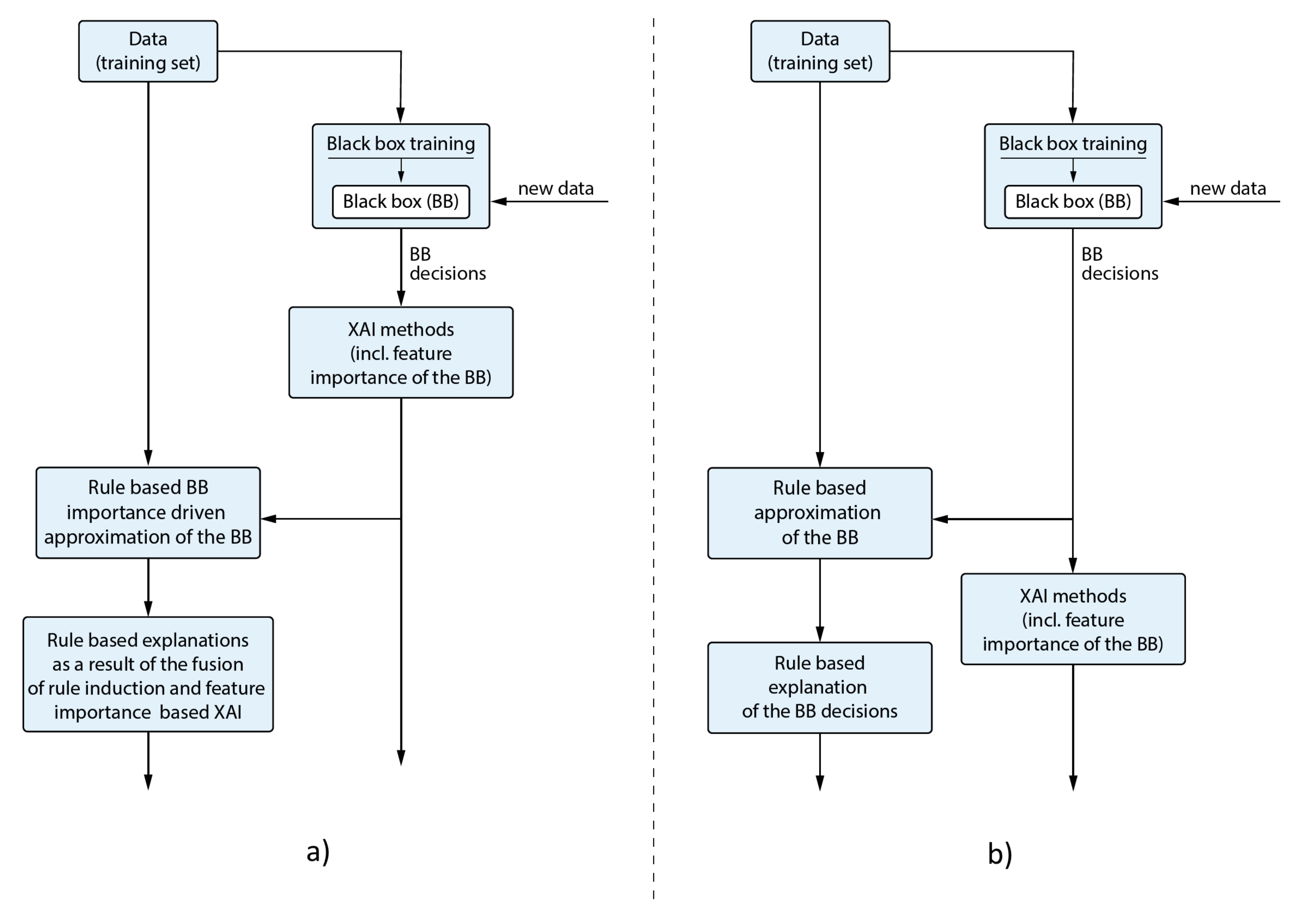}}
    \caption{Illustration of the proposed approach (a) and typical approach (b) for generating rule-based explanations.}
    \label{fig:schema}
\end{figure}

The aim of this study is twofold. Firstly, it aims to verify if the rule-based surrogate models applied to tabular data and classification problems mimic thoroughly the approximated black box model.
Secondly, this study aims to propose a novel method for generating a rule-based white box model, allowing to obtain explanations more consistent with the approximated complex model. Fig. \ref{fig:schema} illustrates the proposed approach allowing a comparison with a typical rule-based explainer generation.
To enable verification of the approach used, software developed as part of the research was published\footnote{ \href{https://github.com/ruleminer/FI-rules4XAI}{https://github.com/ruleminer/FI-rules4XAI}} for other users. In addition, within the software, a detailed report containing the full experimental results on multiple benchmark datasets was made available.

The contributions of this study are as follows.
\begin{itemize}
    \item A comparison of rule-based models approximating the decisions of complex (black box) models with rule-based models trained on the original training data. This comparison verifies whether the generation of an approximator is needed and whether rule-based models generated directly on the training data approximate the black box model equally well as rule-based approximators of these models.
    \item Verification whether the decision bases of the black box model and the rule-based white box model that approximates it are consistent, i.e. whether the same decisions of both models are based on the same, similarly important attributes.
    \item A novel method for generating decision rules to properly explain the black box model decisions, which takes into account the importance of the features of the complex model when generating the rule-based approximator. The proposed method was used to build rule-based factual and counterfactual explanations.
\end{itemize}

The remainder of this paper is organised as follows. Section \ref{sec:related_work} presents the literature review on XAI methods and rule-based XAI methods in particular. Section \ref{sec:materials_and_methods} introduces formal bases for the proposed method, the method itself and the criteria for consistency evaluation of the surogate explaining model and the black box model. The experiments and their results are presented in Section \ref{sec:results}. Section \ref{sec:discussion} discusses the possible pathways for the application of the rule-based approach in XAI and evaluates the rule-based explanations in the context of Co-12 properties. Section \ref{sec:conclusions} concludes the presented study.

\section{Related work}
\label{sec:related_work}

Due to the importance and ubiquity of artificial intelligence and efforts to introduce regulations related to, among others, the explanation of decisions made by AI models, work on XAI methods has intensified significantly in recent years. It resulted in the publication of numerous review papers and books related to this field \cite{BARREDOARRIETA202082,molnar2022,biecek2021explanatory,saeed2023explainable,adadi2018peeking}.

Existing XAI methods can be divided into several classes depending on the perspective adopted for their description. 
One approach is to distinguish methods that can explain the created model at the level of the entire dataset (global methods) or can explain the basis of the model's decisions for a single example (local methods). 
Taking a different perspective, XAI methods can be divided into model specific and model agnostic. In the first case, the explanation method is dedicated to a specific type of model, typically neural network architecture \cite{samek2021explaining}. In the second case, the explanatory method treats the model as a black box and is not related to it in any way \cite{darias2021}. There are many different methods of this type, among which it is possible to distinguish a group of solutions generating an interpretable model approximating the black box model. In this way, both local and global explanations can be generated, and the resulting explanations can take different forms e.g. linear regression coefficients \cite{ribeiro-etal-2016-trust}, decision tree based rules determining counterfactual explanations \cite{guidotti2019factual} or rule-based explantions \cite{MACHA2022101209}. 
Another way of subdividing XAI methods takes into account their application. XAI methods are used in a wide range of applications spanning from medicine \cite{antoniadi2021current,van2022explainable} to industry and predictive maintenance \cite{gama2023xai}.
Moreover, depending on the type of data to be analysed, a distinction can be made between methods applicable to tabular data \cite{sahakyan2021xai}, images \cite{van2022explainable, vermeire2022xai} or text \cite{danilevsky2020}. There are some methods that can be applied to different types of data. The most popular of these are LIME \cite{ribeiro-etal-2016-trust} and SHAP \cite{lundberg2017unified}, and their various modifications \cite{huang2022graphlime, frye2020asymmetric, wang2021shapley, zhou2021s}.

Among the various concepts of how to explain a complex machine learning model, approaches using a rule-based model can be distinguished. Such a model is interpretable, i.e. human-understandable, and can be used to generate both global and local explanations by approximating the complex model. This study focuses on rule-based XAI methods, which are model agnostic treating the model being explained as a black box.

There are many approaches to generating local rule-based explanations. If-then rules called anchors presented in \cite{ribeiro2018anchors} define the regions in the feature space where the decision is (almost) always the same and does not depend on the values of other attributes. 
Another approach defining a local rule with tunable complexity on the basis of a boolean formula is presented in \cite{rosenberg2023explainable}. The classifier presented in this paper that approximates complex model decisions is trained using native local optimisation techniques, efficiently searching the space of feasible formulas.
The method presented in \cite{nossig2023voting} aims to improve the accuracy of the local rule-based explainers by applying weighted voting of multiple rule-based classifiers. The use of an ensemble of rule-based models is expected to better reflect the performance of the complex model being approximated.
Counterfactual explanations based on rules determined from a decision tree are presented in \cite{guidotti2019factual}. The LORE algorithm introduced in this study generates by means of a genetic algorithm a set of examples in the local neighbourhood of the data instance to be explained. Next, it creates a decision tree that is a local approximation of the complex model and derives rules from this tree that are factual and counterfactual explanations. An extension of the method using a stability-enhancing ensemble of explanatory models is presented in \cite{Guidotti_2022}. Another approach in which local (locally optimal) rules are generated for single data instances and which can be used for explanation is presented in \cite{Huynh_2023}. In this approach named LORD, attention was paid not only to the accuracy of the rule but also to its efficiency.

In addition to the methods presented above that generate local explanations, there are many examples of the use of rule-based methods to generate global explanations of complex models. The study \cite{lakkaraju2016interpretable} presents a method for generating decision sets which are sets of independent if-then rules. Decision set learning is performed through an objective function that simultaneously optimises accuracy and interpretability of the rules. This approach should allow efficient approximation and explanation of complex models. 
Another rule-based approach for global explanation of trained models is presented in \cite{ sushil2018rule}. First, the importance of the features in the complex model is calculated in this method. Next, the original inputs with the feature importance scores are weighted and the original input space is transformed for simplification. Finally, a rule-based explanatory model is induced on the derived features.
In \cite{ming2018rulematrix} the RuleMatrix interactive visualization technique applicable to rule-based representation is presented. 
The authors of the \cite{malandri2023model} propose to transform the interpretable (surrogate) model approximating the complex model into a Binary Decision Diagram (BDD) \cite{bryant1992symbolic}, which reflects well (better than, for example, rule sets) the complex relationships between attributes.
The work \cite{cito2021explaining} does not focus on global explanations at the level of the entire dataset, but presents the application of rule induction to generate explanations characterising the regions in data space on which the model performs particularly poorly.
Finally, the RuleXAI library presented in \cite{MACHA2022101209} offers both local and global rule-based explanations. This method makes it possible to determine the importance of individual conditions in the rules comprising the surrogate rule-based model. Thus, it allows a better understanding of which feature, and in which value ranges, contributes to the decision of a complex model.

The presented literature review indicates that, despite the existence of a variety of approaches to generating rule-based explanations, there is a lack of research ensuring that the explanatory rule-based model mimics the behavior of the approximated black box model. This conclusion motivates the research presented.

\section{Materials and Methods}
\label{sec:materials_and_methods}

\subsection{Basic Notions and Definitions}

Let us assume that a universe of possible instances $U$ is given. Each instance $x \in U$ is characterised by a set of attributes $A = \{a_1, \ldots, a_n \}$. Each attribute is interpreted as a function $a: U \rightarrow V_a$, where $V_a$ is a set of possible values.
Two types of attributes are discerned: symbolic (discrete-valued) and numeric (real-valued). 
Each example is also characterised by a special attribute $d$, often called a decision attribute or target. The decision attribute is a discrete class identifier $d: U  \rightarrow V_d$, where $V_d = \{L_1, ..., L_k\}$ and $L_i$, $i\in \{1, ...,k\}$ is a so-called class label or decision class. Furthermore, $d(x)$ denotes the value of the decision attribute for an example $x$. The system $D(X, A, d)$, where $X \subseteq U$ is called a decision table. 
Further notions needed to define the problem of approximating and explaining complex machine learning (black-box) models with a rule-based model are presented in Table \ref{tab:notation}.

\begin{table}[ht]
\caption{Notions used for the definition of approximation and explanation of complex machine learning (black box) models with a rule-based model}
\label{tab:notation}
\centering
\begin{tabular}{lp{0.7\linewidth}}
\hline
Symbol & Description \\ 
\hline
 $M$, $M(x)$  &  the model being explained and its prediction for an instance $x$. \\ 
 $R$, $R(x)$  &  the rule-based model and its prediction for $x$. \\
 $RM$, $RM(x)$ &  the rule-based approximation of $M$ and its prediction for $x$; if $M$ is trained on $(X, A, d)$ then $RM$ is trained on $(X, A, M(X))$. \\
 $F(m)$ &   the set of all features used by the model $m$, where $m \in \{M, R, RM\}$, $F(m) \subseteq A$.\\
 $F(m, x)$ &   the set of features used by the model $m$ to make the prediction for $x$, since some models make predictions based on a subset of features, thus $F(m, x) \subseteq F(m)$. \\
 $fo(m)$ &   importance-based feature ordering for $m$; if $fo(m)=(a_{i_{1}}, ..., a_{i_{s}})$ then $F(m)=\{a_{i_{1}}, ..., a_{i_{s}}\}$.   \\
 $fo(m, x)$ &   importance-based feature ordering for $m$ when the example $x$ is classified; for every $x$, if $fo(m, x)=(a_{i_{1}}, ..., a_{i_{l}})$ then $F(m, x)=\{a_{i_{1}}, ..., a_{i_{l}}\}$. \\
\hline   
\end{tabular}
\end{table}

The rule-based data model $R$ is a set of classification (decision) rules generated by an induction algorithm. Each rule $r \in R$ has the form:
\begin{equation}
\label{eq:rule}
\textrm{IF}\; c_1 \land c_2 \land \ldots \land c_m \; \textrm{THEN}\; d=L.    
\end{equation}

The premise of a rule is a conjunction of conditions $c_j \equiv a_j \odot v_{a_j}$, with $v_{a_j}$ being an element of the attribute $a_j$ domain and $\odot$ representing a relation ($=$ for nominal attributes; $<, \leq, >, \geq$ for numerical ones). 
The conclusion (decision) $d=L$ of the rule indicates the class label $L \in \{L_1,..., L_k\}$. 
If an example $x$ fulfills the conditions of a rule $r$ premise, we say that $r$ covers $x$ ($x$ is covered by $r$).

Given a set of examples $D(X A, d)$ and the rule (\ref{eq:rule}), the set of all examples from $X$ such that $d(x)=L$ is called the set of positive examples, while all other examples are referred to as negative examples. The number of all positive examples and the number of all negative examples are denoted by $P$ and $N$, respectively. The number of positive examples covered by the rule (\ref{eq:rule}) is denoted by $p$, and the number of negative examples covered by (\ref{eq:rule}) is denoted by $n$.

Decision rules are employed both for data description and for classifying new instances with unknown values of the decision attribute. Typically, a rule is interpreted as a dependency that allows one to infer the membership of an example covered by the rule to the decision class indicated in the rule's conclusion. Since, in the presented considerations, rules are induced based on the learning from examples paradigm, this inference can be fallible, which is a feature of inductive inference.

A rule that covers an example $x$ can be perceived as an explanation for the classification of the example to a specific decision class - the class indicated in the conclusion of that rule \cite{ribeiro2018anchors, maszczyk2022rule}. In this case, the abductive reasoning is employed, which, due to the nature of rule induction, can also prove to be fallible.

Among the various approaches to the induction of classification rules \cite{furnkranz1999separate, dembczynski2010, yoon2012class, dash2018boolean, yu2023finrule}, algorithms based on the sequential covering strategy \cite{blaszczynski2011, furnkranz2012foundations, wojtusiak2006, gudys2020rulekit} have demonstrated significant utility in data description and classification of unseen examples.

In this research, sequential covering algorithm is utilised, which efficacy has been confirmed in a number of publications \cite{sikora2013data, GuideR, gudys2020rulekit}. Additionally, an object-related modification of this algorithm dedicated to local explainability is employed. The object-related rule induction approach is motivated by rule induction methods used in Rough Set Theory \cite{sikora2010decision, zhong}, specially by the heuristic strategy of the global and object-related decision reduct finding \cite{hoa1996some}.

In both algorithms, a key role plays the rule quality measure (rule search heuristics (\cite{Furnkranz2005})), which controls the rule induction process. The primary objective of induction is to generate rules that are both precise (covering only positive examples) and general (covering as many examples as possible). 
Therefore, rule quality is typically characterised by both precision and coverage calculated as: 
\begin{equation}
\label{eq:precision}
precision = \frac{p}{p+n} ,
\end{equation}
\begin{equation}
\label{eq:coverage}
coverage = \frac{p}{P} .
\end{equation} 
There is a wide range of rule quality measures that are used to supervise the rule induction process and rule assessment \cite{furnkranz2012foundations, lavravc1999rule, bruha2003rule, greco2004can} presented in literature. Depending on the measure employed, different classification models are obtained. Optimal results are usually achieved using measures that consider both the precision and coverage of a rule, and the values of P and N \cite{janssen2010quest, wrobel2016rule}. In this study, in addition to the above measures, measure C2, which has the following form, was used to generate rules:
\begin{equation}
\label{eq:c2}
C2 = \left( \frac{Np - Pn}{N(p+n)} \right)\left( \frac{P + p}{2p} \right) .
\end{equation}

\subsection{Fusion of rule-based explanations with importance-based feature ordering}

The idea of the sequential covering approach following the separate-and-conquer heuristic is presented in (Algorithm~\ref{alg:conquer}). Rule generation is carried out sequentially for each class and its positive examples. The rules are added iteratively to the initially empty set as long as the entire dataset becomes covered. Every rule must cover at least $mincov$ previously uncovered examples, to ensure the convergence. Therefore, when there are less then $mincov$ uncovered examples left, the generation of consecutive rules is stopped. The induction of a single rule consists of two stages: growing and pruning. In the growing stage (presented in Algorithm~\ref{alg:grow}), elementary conditions are added to the initially empty premise. All possible conditions built on all attributes (line 6: \textsc{GetPossibleConditions} function call) are considered in this step. The conditions leading to the rule with the best quality are selected (lines 10--12). 
In the case of nominal attributes, conditions in the form $a_j = v_j$ for all values $v_j$ from the attribute $a_j$ domain are considered. For continuous attributes, $a_j$ values that appear in the observations covered by the rule are sorted. Then, the possible split points $v_j$ are determined as arithmetic means of subsequent $a_j$ values and conditions $a_j < v_j$ and $a_j \geq v_j$ are evaluated. If several conditions lead to the same results, the one covering more examples is chosen. 
In the pruning stage, conditions are iteratively removed from the rule premise to achieve the greatest improvement in rule quality. The procedure stops when no conditions can be deleted without decreasing the quality of the rule or when the rule contains only one condition. Finally, the comprehensibility of the rule is improved by merging conditions based on the same numerical attributes. A broader general outline of the sequential covering algorithm is discussed, among others, in \cite{furnkranz2012foundations,gudys2020rulekit}.

\begin{algorithm}[!htb]
	\small
		\caption{Separate-and-conquer rule induction.}
		\label{alg:conquer}
	\begin{algorithmic}[1]		
		\Require $D(X, A, d)$---training dataset,
		$mincov$---minimum number of yet uncovered positive examples that a new rule has to cover.
		\Ensure $R$---rule set.
		\State $R \gets \emptyset$ \Comment{start from an empty rule set}
    \For{$L \in \{L_1,...,L_k\}$}    \Comment{for each decision class}
        \State $Pos \gets D(X, A, L)$   \Comment{set of positive examples from $D$}
		\Repeat
    		\State $r \gets \emptyset$, $d=L$ \Comment{start from a rule with empty premise and conclusion $d=L$}
    		\State $r \gets \Call{Grow}{r, D, Pos, mincov}$ \Comment{grow the rule}
    		\State $r \gets \Call{Prune}{r,D, Pos}$ \Comment{prune the rule}
    		\State $R \gets R \cup \{r\}$
    		\State $Pos \gets Pos\setminus\Call{Cov}{r, Pos}$ \Comment{remove from $Pos$ examples covered by $r$}
		\Until{$|Pos| > 0$}
    \EndFor
	\end{algorithmic}
\end{algorithm}

\begin{algorithm}[!htb]
		\caption{Growing a rule.}
		\label{alg:grow}
	\begin{algorithmic}[1]
		\Require
		$r$---input rule,
		$D$---training dataset,
		$D_{U}$---set of uncovered examples,
		$mincov$---minimum number of previously uncovered examples that a new rule has to cover.
		\Ensure
		$r$---grown rule.
		
		\Function{Grow}{$r$, $D$, $D_U$, $mincov$}
		
		\Repeat \Comment{iteratively add conditions}
		\State $c_\textrm{best} \gets \emptyset$ \Comment{current best condition}
		\State $q_\textrm{best} \gets -\infty,\quad \textrm{cov}_\textrm{best} \gets -\infty$ \Comment{best quality and coverage}
		
		\State $D_{r} \gets$ \Call{Cov}{$r$, $D$} \Comment{examples from $D$ satisfying $r$ premise}
		
		\For{$c \in$ \Call{GetPossibleConditions}{$D_r$}}
		\State $r_c \gets r \land c$ \Comment{rule extended with condition $c$}
		\State $D_{r_c} \gets \Call{Cov}{r_c, D}$
		\If {$|D_{r_c} \cap D_U| \geq mincov$} \Comment{verify coverage requirement}
		\State $q \gets$ \Call{Quality}{$r, D$} \Comment{rule quality measure}
		
		\If {$q > q_\textrm{best}$ \textbf{or} ($q = q_\textrm{best}$ \textbf{and} $|D_{r_c}| > \textrm{cov}_\textrm{best}$)}
		\State $c_\textrm{best} \gets c,\quad q_\textrm{best} \gets q$,\quad $\textrm{cov}_\textrm{best} \gets |D_{r_c}|$
		\EndIf
		
		\EndIf
		\EndFor			
		\State $r \gets r \land c_\textrm{best}$
		\Until{$c_\textrm{best} = \emptyset$}
		\State \Return{$r$}
		\EndFunction
	\end{algorithmic}
\end{algorithm}

The separate-and-conquer approach (Algorithm \ref{alg:conquer}) generates unordered set of rules. 
Therefore, the classification of a data example, e.g. a test one, requires the evaluation of a set $R_\textrm{cov} \subseteq R$ of rules that cover the example and the aggregation of the results obtained. This differs from ordered rule sets (decision lists), where the first rule covering the investigated observation determines a model response. Aggregation, in the case of classification, is achieved using voting. Each rule from $R_\textrm{cov}$ votes with its value of the quality measure.

The aim of this study is to modify the sequential covering approach to ensure that the rule-based explanations are based on the attributes that are important for the decisions made by the approximated complex model.
Therefore, the algorithm presented in Algorithm \ref{alg:conquer} and \ref{alg:grow} was subjected to the following modifications. The proposed approach assumes that the importance-based feature ordering ($fo(M)$ or $fo(M,x)$) determined for the black box model $M$ is passed to the algorithm. This ranking specifies a sequence of attributes which may be a subset of the set of all attributes.
Furthermore, the proposed approach was based on the object-related rule induction. In this approach one rule is generated to cover (to explain) a specific example.
Specifically, for a given set of examples (e.g., a training set), one rule is generated for each data instance. 
The pseudo-code of the proposed approach modifying Algorithm \ref{alg:conquer} is presented in Algorithm \ref{alg:conquer_M2}. A new rule is generated for each class $L$ and for each data example $x$ belonging to this class (line 4). Rule induction is performed using the importance-based ordering of the attributes, which is denoted generally as $fo$ in the Algorithm \ref{alg:conquer_M2}. The ordering can be global, i.e. identical for each data example (e.g. it can be $fo(M)$ resulting from the global explanation of the black box model $M$). It can be local, i.e. different for each example $x$ on which the rule is generated (e.g. it can be $fo(M,x)$ resulting from the local explanation of the black box model $M$ decision for data example $x$).

Furthermore, the use of importance-based ordering of the attributes imposes a change in the \textsc{Grow} method implementing the rule growth. The pseudo-code of the proposed modification of Algorithm \ref{alg:grow} is presented in Algorithm \ref{alg:grow_M2}.
Since a rule is generated within the object-related rule induction on a single data example, line 5 of the \textsc{Grow} method must be changed to the form presented in \ref{alg:grow_M2}. 
Further changes result from the use of importance-based feature ordering ($fo$).
In this approach, the set of attributes $Ao$ is introduced in line 7 of Algorithm \ref{alg:grow_M2}. Initially, $Ao$ contains only the most important attribute, then successively more attributes are added according to the $fo$ ranking. In each iteration, $Ao$ is passed to the \textsc{GetPossibleConditions} function (line 8).
Thus, in the proposed modification, unlike in the approach presented in Algorithm \ref{alg:grow}, not all attributes describing the training data are passed to the \textsc{GetPossibleConditions} function but iteratively expanding subset $Ao$ of the attributes.  
Operation of the \textsc{GetPossibleConditions} function remains unchanged. It generates the rule conditions for all attribute values or possible split points depending on the nominal or numerical type of the attribute.
The pruning stage remains unchanged in this approach.

The Algorithm \ref{alg:conquer_M2} is completed with optional filtering of the resulting rule set to reduce the number of rules. Filtering is performed on a set of rules that are sorted in descending order according to the quality measure. Iteratively for each successive rule, those examples that are covered by the rule are removed from the set of examples. Filtering is completed when all examples are covered (removed).

\begin{algorithm}[!htb]
	\small
		\caption{Object-related rule induction.}
		\label{alg:conquer_M2}
	\begin{algorithmic}[1]		
		\Require $D(X, A, d)$---training dataset,
		$mincov$--- a minimum number of yet uncovered positive examples that a new rule has to cover,
        $fo$---importance-based feature ordering,
        $filtering$---flag indicating whether to filter the rules.
		\Ensure $R$---rule set.
		\State $R \gets \emptyset$ \Comment{start from an empty rule set}
    \For{$L \in \{L_1,...,L_k\}$}    \Comment{for each decision class}
        \State $Pos \gets D(X, A, L)$   \Comment{set of positive examples from $D$}
		\For{$x \in Pos$} \Comment{for each positive example}
    		\State $r \gets \emptyset$, $d=L$ \Comment{start from a rule with an empty premise and conclusion $d=L$}
    		\State $r \gets \Call{Grow}{r, x, D, Pos, mincov, fo}$ \Comment{grow the rule}
    		\State $r \gets \Call{Prune}{r,D, Pos}$ \Comment{prune the rule}
    		\State $R \gets R \cup \{r\}$
		\EndFor
    \EndFor
    \If{$filtering$}
    \State $R \gets \Call{Filtering}{R}$    \Comment{filter the rules}
    \EndIf
	\end{algorithmic}
\end{algorithm}

\begin{algorithm}[!htb]
		\caption{Attribute importance-driven rule growth}
		\label{alg:grow_M2}
	\begin{algorithmic}[1]
		\Require
		$r$---input rule,
		$D$---training dataset,
		$D_{U}$---set of uncovered examples,
		$mincov$---minimum number of previously uncovered examples that a new rule has to cover,
        $fo=(a_{i_{1}}, ..., a_{i_{s}})$ ---importance-based feature ordering.
		\Ensure
		$r$---grown rule.
		
		\Function{Grow}{$r$, $D$, $D_U$, $mincov$, $fo$}
		
		\Repeat \Comment{iteratively add conditions according to the $fo$ ordering}
		\State $c_\textrm{best} \gets \emptyset$ \Comment{current best condition}
		\State $q_\textrm{best} \gets -\infty,\quad \textrm{cov}_\textrm{best} \gets -\infty$ \Comment{best quality and coverage}
		
		\State $D_{r} \gets x$  \Comment{example for which the rule is generated}

  \For{$j \in \{1, \ldots, s \}$}    \Comment{for consecutive attributes in $fo$ ordering}
        \State $Ao \gets \{ a_{i_1}, \ldots, a_{i_j}\} $  \Comment{select subset of the most important attributes}
		\For{$c \in$ \Call{GetPossibleConditions}{$D_r, Ao$}}
		\State $r_c \gets r \land c$ \Comment{rule extended with condition $c$}
		\State $D_{r_c} \gets \Call{Cov}{r_c, D}$
		\If {$|D_{r_c} \cap D_U| \geq mincov$} \Comment{verify coverage requirement}
		\State $q \gets$ \Call{Quality}{$r, D$} \Comment{rule quality measure}
		
		\If {$q > q_\textrm{best}$ \textbf{or} ($q = q_\textrm{best}$ \textbf{and} $|D_{r_c}| > \textrm{cov}_\textrm{best}$)}
		\State $c_\textrm{best} \gets c,\quad q_\textrm{best} \gets q$,\quad $\textrm{cov}_\textrm{best} \gets |D_{r_c}|$
		\EndIf		
		\EndIf
		\EndFor			
  \EndFor	
		\State $r \gets r \land c_\textrm{best}$
		\Until{$c_\textrm{best} = \emptyset$}
		\State \Return{$r$}
		\EndFunction
	\end{algorithmic}
\end{algorithm}

\subsection{Evaluation of rule-based explanations}
\label{sec:evaluation_conditions}

As outlined in the Introduction section, the aim of this study is to verify the adequacy of the explanations of a black box model $M$ generated by a rule-based model $R$ (especially the explanations generated by a rule-based approximation of the black box model -- $RM$). Therefore, the following three conditions were defined for $R$ that can be used to explain $M$:
\begin{equation}\label{eq:cond1}
        \forall X \subseteq U, \ \forall x \in X \ \ R(x) = M(x),
\end{equation}
which means that the explainer $R$ and the model being explained $M$ make the same decisions on each data set,
\begin{equation}\label{eq:cond2}
     F(R) = F(M),
\end{equation}
which means that the explainer $R$ and the model being explained $M$ make decisions on the basis of the same features,
    \begin{equation}\label{eq:cond3}
        \forall X \subseteq U, \ \forall x \in X \ \ fo(R,x) = fo(M,x),
    \end{equation}
which means that the importance-based ordering (ranking) of the features on which the explainer $R$ and the model being explained $M$ make decisions is the same.
An ideal explanatory model is one that meets all the conditions listed above. 

With respect to defined conditions (\ref{eq:cond1}-\ref{eq:cond3}), the fidelity of the explainer $R$ can be calculated using various methods.
Criterion (\ref{eq:cond1}) can be verified using a measure of agreement between the decisions of two classifiers, e.g., by determining the Cohen kappa coefficient between models $R$ and $M$. Criterion (\ref{eq:cond2}) can be verified by measuring the mutual inclusion between the feature sets used by the models $R$ and $M$: 
\begin{equation}\label{eq:mutual}
        \nu = \frac{|F(R) \cap F(M)|}{|F(R) \cup F(M)|} .
\end{equation}
Criterion (\ref{eq:cond3}) can be verified by comparing the importance rankings of attributes using, e.g. by determining Kendall's Tau correlation coefficient.

\section{Experiments and results}
\label{sec:results}

The proposed method was extensively verified on numerous datasets in the experiments carried out. The conditions presented in Section \ref{sec:evaluation_conditions}, which define the ideal explanatory model, determined the scheme of the analyses conducted. The measures indicated there to verify the quality of the proposed model were used to evaluate and compare the quality of the proposed rule model generation method for black box model explanations. 

The analyses carried out were divided into three parts. 
In the first, condition (\ref{eq:cond1}) of Section  \ref{sec:evaluation_conditions} stating that the generated approximator should make the same decisions as the approximated black box model was verified. The second part verified the consistency of the basis for the decisions made by the model, i.e. the explanations showing on which attributes the rule-based approximator and the black box model made decisions. This analysis includes condition (\ref{eq:cond2}) and condition (\ref{eq:cond3}), which refer to the overlap of important attributes for each model and the consistency of the ranking of these attributes. The final part of the analyses conducted was a comparison of the proposed method with the Anchors method \cite{ribeiro2018anchors}, selected as the state-of-the-art rule-based XAI solution. The general comparison carried out was complemented by a case study on selected data for both methods. The remainder of this section presents the aggregated results of the experiments and the conclusions drawn from them. A detailed report containing the full experimental results and the software used in the experiments are available online (\href{https://github.com/ruleminer/FI-rules4XAI}{https://github.com/ruleminer/FI-rules4XAI}).

\subsection{Datasets and experimental settings}

The proposed approach was verified on 30 datasets from UCI\footnote{https://archive.ics.uci.edu/} and OpenML\footnote{https://www.openml.org/} repositories. 
Datasets dedicated to the classification task were selected and care was taken to ensure their diversity. The dataset names and characteristics (number of observations, attributes, and classes) are presented in Table \ref{tab:datasets}. 

\begin{table}[!htb]
\caption{Datasets (abbreviated dataset names: *kdd-synthetic-control, $^\dagger$mammographic-masses, $^\ddagger$seismic-bumps, $^\mathsection$wall-robot-navigation) and their characteristics - number of observations, attributes and classes}
\label{tab:datasets}
\centering
\begin{tabular}{lccclccc}
\hline
Dataset & $|D|$ & $|A|$ & $|V_d|$ & Dataset & $|D|$ & $|A|$ & $|V_d|$ \\ 
\hline
cylinder-bands	&	540	&	35	&	2	&	titanic	&	2201	&	3	&	2	\\
wdbc	&	569	&	30	&	2	&	segment	&	2310	&	19	&	7	\\
kdd*	&	600	&	60	&	6	&	seismic$^\ddagger$	&	2584	&	18	&	2	\\
balance-scale	&	625	&	4	&	3	&	madelon	&	2600	&	500	&	2	\\
soybean	&	683	&	35	&	19	&	dna	&	3186	&	180	&	3	\\
credit-a	&	690	&	15	&	2	&	splice	&	3190	&	60	&	3	\\
breast-w	&	699	&	9	&	2	&	kr-vs-kp	&	3196	&	36	&	2	\\
diabetes	&	768	&	8	&	2	&	wilt	&	4839	&	5	&	2	\\
vehicle	&	846	&	18	&	4	&	churn	&	5000	&	20	&	2	\\
tic-tac-toe	&	958	&	9	&	2	&	phoneme	&	5404	&	5	&	2	\\
mammo$^\dagger$	&	961	&	5	&	2	&	wall-robot$^\mathsection$	&	5456	&	24	&	4	\\
qsar-biodeg	&	1055	&	41	&	2	&	mushroom	&	8124	&	22	&	2	\\
semeion	&	1593	&	256	&	10	&	har	&	10299	&	561	&	6	\\
car	&	1728	&	6	&	4	&	nursery	&	12960	&	8	&	5	\\
cmc	&	2000	&	47	&	10	&	nomao	&	34465	&	118	&	2	\\
\hline
\end{tabular}
\end{table}

Moreover the proposed method was applied to approximate and explain three different types of black box models generated by the following methods: extreme gradient boosting (XGB), support vector machines (SVM) and random forest (RF). 
Raw data were preprocessed by applying one hot encoding of categorical attributes and imputation of missing values. Each dataset was split into train and test subsets in 80/20 proportion. The models were evaluated using average balanced accuracy. Exhaustive evaluation, e.g. using cross-validation, and high quality of the models was not the most important given the aim of the work which was to generate explanations.
Data processing and model training were performed using the scikit-learn Python package \cite{scikit-learn}. 

\begin{figure}[!htb]
    \centerline{\includegraphics[width=0.9\textwidth]{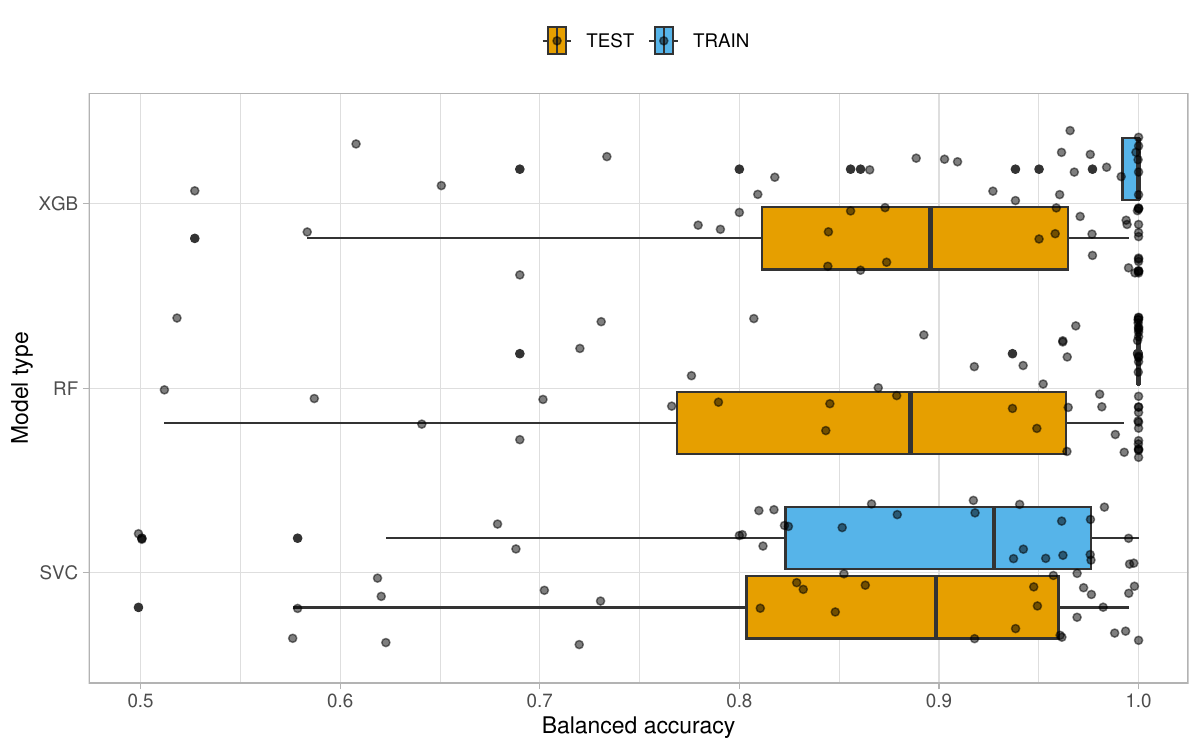}}
    \caption{Performance of the generated black box models expressed as average balanced accuracy}
    \label{fig:bacc}
\end{figure}

Performance of the black box models on training and test data is illustrated in Fig. \ref{fig:bacc}.
The resulting average balanced accuracy values show that the generated models have a high predictive quality.

\subsection{Evaluation of rule-based approximation}

The first part of the experiments aimed to evaluate the quality of the rule-based approximation.
Two approaches to approximation using a rule-based model were verified. In the first, the rule-based model is generated on the original data on which the black box model was generated. In this case, the quality of the model describing the analysed phenomenon is assessed. In the second approach, the rule-based model is generated on data in which the decision column contains the decisions of the black box model. In this case, the quality of the model that approximates the decisions of the black box model is assessed.
Moreover, two methods of the rule-based model generation were verified. The RuleKit method \cite{gudys2020rulekit} was used as the implementation of standard separate-and-conquer approach and is hereafter referred to as SC-S (separate-and-conquer - standard). The RMatrix method \cite{GuideR} was used as the implementation of object-related approach and is hereafter referred to as SC-OR (separate-and-conquer - object-related). 
In each method, two measures of quality assessment were verified in the rule growth process: precision and C2.

To assess the quality of approximation the Cohen's kappa coefficient between black box model predictions and rule-based model predictions was calculated on train and test data.
The results for the approach on the original data and on the black box model decision data are presented in Table 
\ref{tab:approx_quality_mean_kappa}.

The obtained results show that the approximation quality
of the rule-based models generated on the dataset with black box model decisions are better than the models generated on original data. The mean kappa values are greater than or the same for this data representation in all cases except two. Therefore, further comparisons will be made for models generated on the dataset with black box model decisions.
The difference between the methods of rule-based model generation and the quality measure used is particularly apparent for the training data. The results obtained on them show that the precision measure allows to obtain a model with an overall excellent approximation quality and clearly better than the C2 measure.
Finally, the results show that the rule-based approximator generated by the SC-OR method was superior.

\begin{table}[!htb]
\caption{Quality of approximation of the black box model ($M$) by the rule-based model ($RM$) expressed by the mean value of Cohen's kappa coefficient. Rule-based models generated on original data and data with black box model decisions were verified. Precision and C2 rule-quality measures were used in rule growth procedure.}
\label{tab:approx_quality_mean_kappa}
\centering
\begin{tabular}{llcccc|cccc}
\hline
$M$	&	$RM$&	\multicolumn{4}{c}{Original data} &	\multicolumn{4}{c}{Black box model decisions}	\\
	&		&	\multicolumn{2}{c}{Precision} &	\multicolumn{2}{c}{C2}	&	\multicolumn{2}{c}{Precision} &	\multicolumn{2}{c}{C2} \\
 	&		&	Train	&	Test	&	Train	&	Test	&	Train	&	Test	&	Train	&	Test	\\
\hline
RF	&	SC-OR	&	0.99	&	0.84	&	0.87	&	0.82	&	0.99	&	0.83	&	0.91	&	0.82	\\
RF	&	SC-S	&	0.93	&	0.81	&	0.88	&	0.83	&	0.95	&	0.82	&	0.89	&	0.84	\\
SVM	&	SC-OR	&	0.8	&	0.76	&	0.79	&	0.77	&	1	&	0.83	&	0.95	&	0.84	\\
SVM	&	SC-S	&	0.78	&	0.72	&	0.78	&	0.75	&	0.98	&	0.84	&	0.93	&	0.84	\\
XGB	&	SC-OR	&	0.97	&	0.84	&	0.88	&	0.82	&	1	&	0.82	&	0.92	&	0.82	\\
XGB	&	SC-S	&	0.93	&	0.82	&	0.89	&	0.82	&	0.96	&	0.83	&	0.9	&	0.82	\\
\hline
\end{tabular}
\end{table}

The mean kappa values presented in Table \ref{tab:approx_quality_mean_kappa} allowed the selection of the rule-based model generation method used in further experiments.
Therefore, the implementation of the proposed method was based on the SC-OR approach and the precision measure was used in the rule growth procedure. The rule-based model was generated on data in which the decision column contained the decisions of the black box model.

\subsection{Evaluation of explanations}
\label{sec:evaluation_explanations}

The second stage of the experiments was dedicated to verifying how well the generated rule-based explanations mimic the black box model decision making.
For this purpose, the consistency of the generated rules with the black box model at the attribute level was verified. Two measures were chosen to assess such consistency, as suggested in Section \ref{sec:evaluation_conditions}. Inclusion was chosen to compare the sets of most important attributes for black box model and its approximator. It shows whether the same key features are used by both models. Inclusion was determined as mutual inclusion (see formula (\ref{eq:mutual})). Correlation was chosen to compare the rankings of the most important features for black box model and its approximator. It shows whether the same most important features rank in the same order for each model. Correlation was determined as Kendall's Tau correlation coefficient. For both measures, unique features occurring in the elementary conditions of the rule explaining the data instance were selected for comparison. The number of these features determined the number of the most important features of the black box model that were taken for comparison.

The experiments used rule-based models generated by the SC-OR method. These models consist of rules generated for each data example. 
Three models generated according to the approaches described in Section \ref{sec:materials_and_methods} were compared, and the induction of the rules forming these models used: (i) global importance-based feature ordering common to all rules, (ii) local importance-based feature ordering determined for each data instance separately, (iii) a basic approach that does not use any feature ranking.
The global feature ranking was determined for the black box model using an approach referred to as Permutation Feature Importance. The local ranking was determined using the SHAP method, which was applied to each black box model decision (each data instance). Unfortunately, generating SHAP-based explanations for the SVM model (to which TreeSHAP is not applicable) was too time-consuming and no results were obtained for this model.

The resulting values representing the consistency of the generated rules with the black box model are presented in Table \ref{tab:consistency}. It presents the aggregated results for experiments conducted on 30 datasets, which include distributions of the Inclusion and Correlation values calculated for black box models and their rule-based explainers. For example, if the Model is specified as RF and the Feature importance is specified as Global, this row contains the distribution of Inclusion and Correlation values determined for models obtained using the Random Forest method and their rule-based explainers, which use information about the global importance-based feature ordering derived for Random Forest. More precisely, these values describe consistency of the features used in rule premises and the global feature ranking derived for Random Forest. If the Feature importance is specified as Local, then the local importance-based feature ordering is derived for each Random Forest decision and these rankings are used in rule generation process. Finally, if the rules are generated without the use of feature importance, then the features from the rule premises are compared with the features indicated as the most important in the global importance-based feature ordering derived for Random Forest.

\begin{figure}[!htb]
    \centerline{\includegraphics[width=0.9\textwidth]{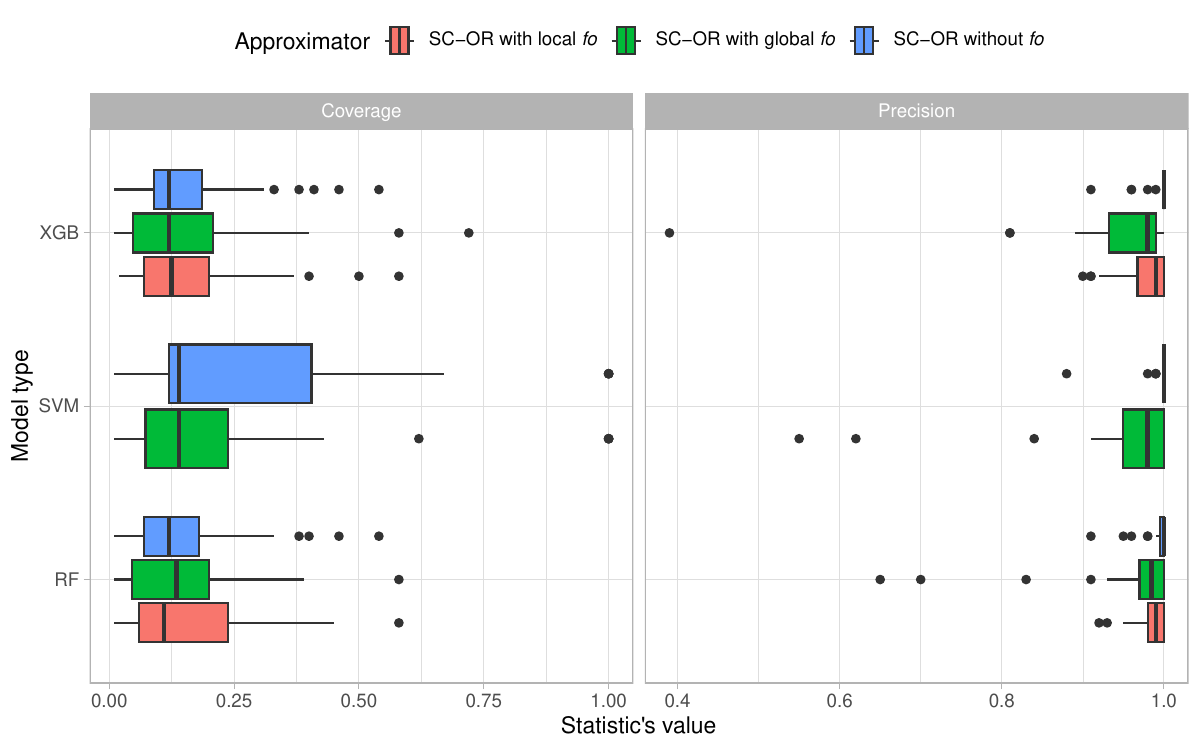}}
    \caption{Quality (precision and coverage) of the rules generated by the approximator that used local, global, and none feature ordering}
    \label{fig:rule_quality}
\end{figure}

The results show that the use of the proposed rule-based methods utilizing both global and local importance-based feature ordering clearly improves the consistency of the rule-based approximator with the black box model. Both Inclusion and Correlation values are higher and closer to 1 when feature ranking was used for rule generation in the SC-OR method.
The quality of the rules generated in this experiment is presented in Fig. \ref{fig:rule_quality}. The results presented in the charts show that the rules generated using the importance-based feature ordering are characterized by high but slightly lower precision (the decrease is greater for the global ranking) and predominantly no worse coverage.

\begin{table}[!htb]
\caption{Consistency of the generated rules with the black box model ($M$) for different variants of the proposed SC-OR approach. Consistency is expressed as Inclusion (mutual inclusion) and Correlation (Kendall's Tau) values for the most important attributes. Variants of the SC-OR method include approaches using global and local importance-based feature ordering and a basic approach without feature importance.}
\label{tab:consistency}
\centering
\begin{tabular}{llcccc|cccc}
\hline
    $M$	&	SC-OR &	\multicolumn{4}{c}{Inclusion}		&	\multicolumn{4}{c}{Correlation} \\
    &	&	Q1	&	Mean	&	Median	&	Q3	&	Q1	&	Mean	&	Median	&	Q3	\\
\hline
RF	&	Global	&	0.33	&	0.68	&	0.71	&	1	&	0	&	0.68	&	1	&	1	\\
RF	&	Local	&	0.33	&	0.69	&	0.71	&	1	&	0	&	0.69	&	1	&	1	\\
RF	&	Without	&	0	&	0.18	&	0	&	0.33	&	0	&	0.11	&	0	&	1	\\
SVM	&	Global	&	0.5	&	0.71	&	0.78	&	1	&	0.33	&	0.72	&	1	&	1	\\
SVM	&	Without	&	0	&	0.25	&	0	&	0.33	&	0	&	0.15	&	0	&	1	\\
XGB	&	Global	&	0.5	&	0.74	&	1	&	1	&	1	&	0.78	&	1	&	1	\\
XGB	&	Local	&	0.5	&	0.76	&	1	&	1	&	1	&	0.76	&	1	&	1	\\
XGB	&	Without	&	0	&	0.24	&	0.11	&	0.33	&	0	&	0.15	&	0	&	1	\\
\hline
\end{tabular}
\end{table}

\begin{figure}[!htb]
    \centerline{\includegraphics[width=0.9\textwidth]{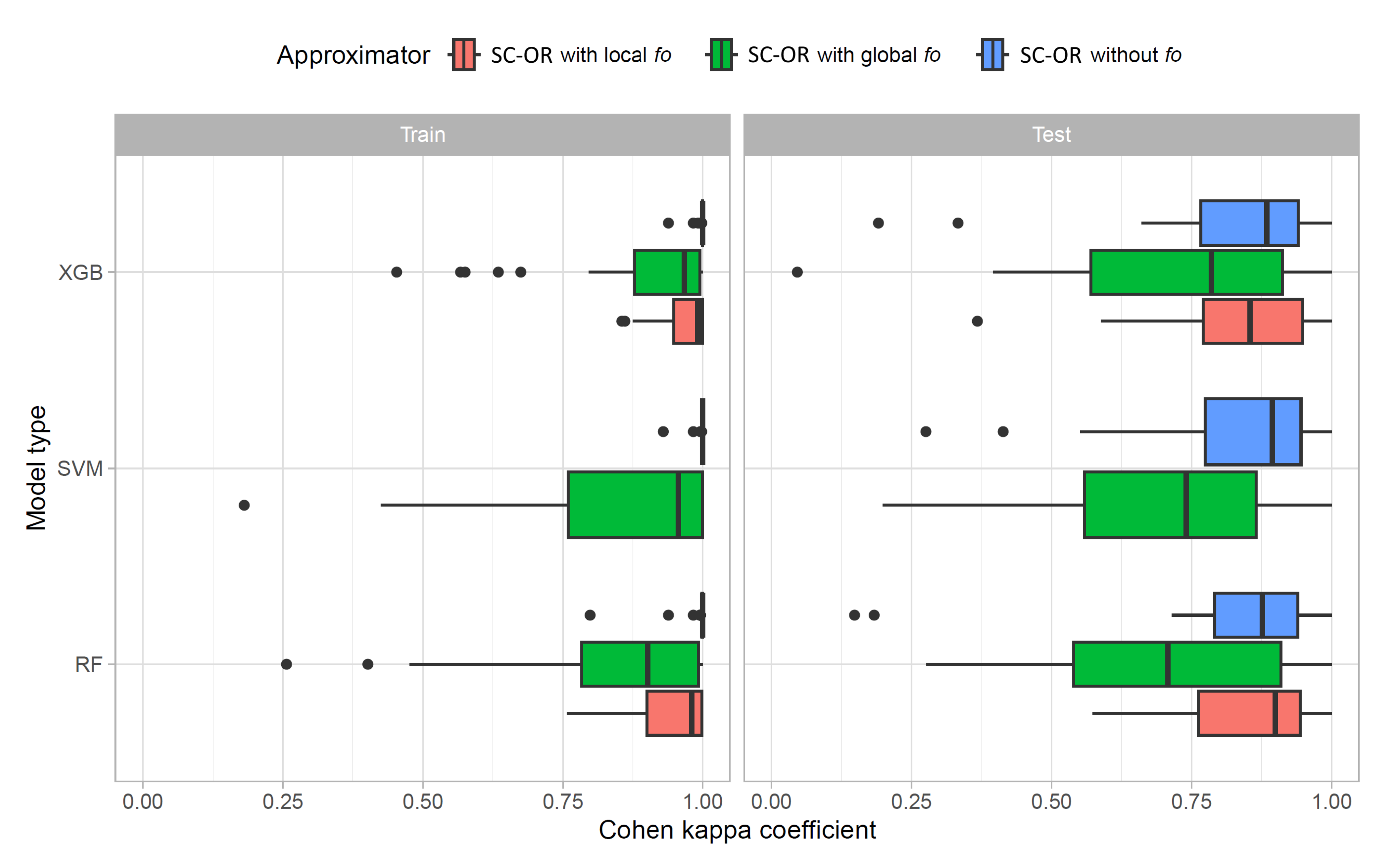}}
    \caption{Approximation quality of the proposed importance-driven rule growth on the training data (left) and test data (right) - Cohen's kappa coefficients between the black box model predictions and rule-based approximations}
    \label{fig:approx_fo}
\end{figure}

In addition, it was verified how the use of importance-based feature ordering affects the quality of the approximation. Approximation quality expressed Cohen's Kappa values calculated for the rule-based models evaluated in Table \ref{tab:consistency} is presented in Fig. \ref{fig:approx_fo}. The plots in Fig. \ref{fig:approx_fo} show that the use of importance-based feature ordering can reduce the quality of the approximation. In the case of local ranking, the change is small; in the case of global ranking, the difference in quality is evident. 

In a further experiment the proposed solution implemented in the SC-OR method was compared with the reference Anchors method \cite{ribeiro2018anchors}. The concept of the experiments was analogous to the approach presented above. However, instead of an approach that does not take into account attribute ordering, the Inclusion and Correlation values for the Anchors algorithm were determined. 
Unfortunately, when using the Anchors algorithm for larger datasets, the generation of explanations for each data instance was aborted due to excessive analysis time. The explanation of a single decision by the Anchors algorithm implementation\footnote{https://github.com/SeldonIO/alibi/} used is approximately twice as long as the SC-OR implementation, and moreover, this implementation does not perform the analysis in parallel. 
Therefore, the analysis was carried out for the following 21 datasets: balance-scale, breast-w, car, churn, cmc, credit-a, cylinder-bands, diabetes, kdd-synthetic-control, kr-vs-kp, mammographic-masses, mushroom, nursery, phoneme, qsar-biodeg, segment, tic-tac-toe, titanic, vehicle, wall-robot-navigation, wdbc. 
Besides, the methods used and settings adopted were the same as in the experiments in Section \ref{sec:evaluation_explanations}.

The calculated consistency of the rule-based approximators with the black box model is presented in Table \ref{tab:anchor_consistency}. 
The results presented in this table follow a format analogous to the results presented for the previous experiment. If Feature importance is specified as SC-OR Global, this means that the consistency values in the table refer to the global feature ranking calculated for the black box model and the features used in the rule premise when the rules were generated by the SC-OR method using this global importance-based feature ordering.
If Feature importance is specified as Anchors Global, this means that the consistency values in the table refer to the global feature ranking calculated for the black box model and the features used in the premise of the rules generated by Anchors. 

The results clearly show that the proposed SC-OR solution achieves its goal generating a rule-based approximator that is more consistent with the black box model and its attribute ranking. The Inclusion and especially Correlation values obtained show how different the rule-based models (and explanations) can be when one of them is able to adjust to the attribute importance determined for the black box model.

\begin{table}[!htb]
\caption{Consistency of the rules (generated by model $RM$) with the black box model ($M$) for the proposed solution (SC-OR) and the Anchors algorithm. Consistency is expressed by Inclusion (mutual inclusion) and Correlation (Kendall's Tau) values for the most important attributes. Variants of the SC-OR method include approaches using global and local importance-based feature ordering, the rules generated by Anchors were induced without any feature ranking.}
\label{tab:anchor_consistency}
\centering
\begin{tabular}{lp{0.1\linewidth}cccc|cccc}
\hline
	$M$    &		$RM$  &	\multicolumn{4}{c}{Inclusion}		&	\multicolumn{4}{c}{Correlation}   \\
	&  &	Q1	&	Mean	&	Median	&	Q3	&	Q1	&	Mean	&	Median	&	Q3	\\
\hline														
RF	&	SC-OR Global	&	0.43	&	0.72	&	1	&	1	&	0	&	0.63	&	1	&	1	\\
RF	&	Anchors Global	&	0.2	&	0.44	&	0.33	&	0.6	&	0	&	0.29	&	0.33	&	0.67	\\
RF	&	SC-OR Local	&	0.5	&	0.76	&	1	&	1	&	0	&	0.65	&	1	&	1	\\
RF	&	Anchors Local	&	0.2	&	0.42	&	0.33	&	0.6	&	0	&	0.19	&	0.24	&	0.67	\\
SVM	&	SC-OR Global	&	0.5	&	0.77	&	1	&	1	&	0	&	0.6	&	1	&	1	\\
SVM	&	Anchors	Global &	0.11	&	0.41	&	0.33	&	0.67	&	0	&	0.24	&	0.08	&	0.6	\\
XGB	&	SC-OR Global	&	0.5	&	0.75	&	1	&	1	&	0	&	0.67	&	1	&	1	\\
XGB	&	Anchors	Global &	0.3	&	0.52	&	0.33	&	1	&	0	&	0.4	&	0.33	&	1	\\
XGB	&	SC-OR Local	&	0.5	&	0.78	&	1	&	1	&	0	&	0.66	&	1	&	1	\\
XGB	&	Anchors	Local &	0.2	&	0.34	&	0.33	&	0.5	&	0	&	0.32	&	0.33	&	1	\\
\hline
\end{tabular}
\end{table}

\begin{figure}[!htb]
    \centerline{\includegraphics[width=0.9\textwidth]{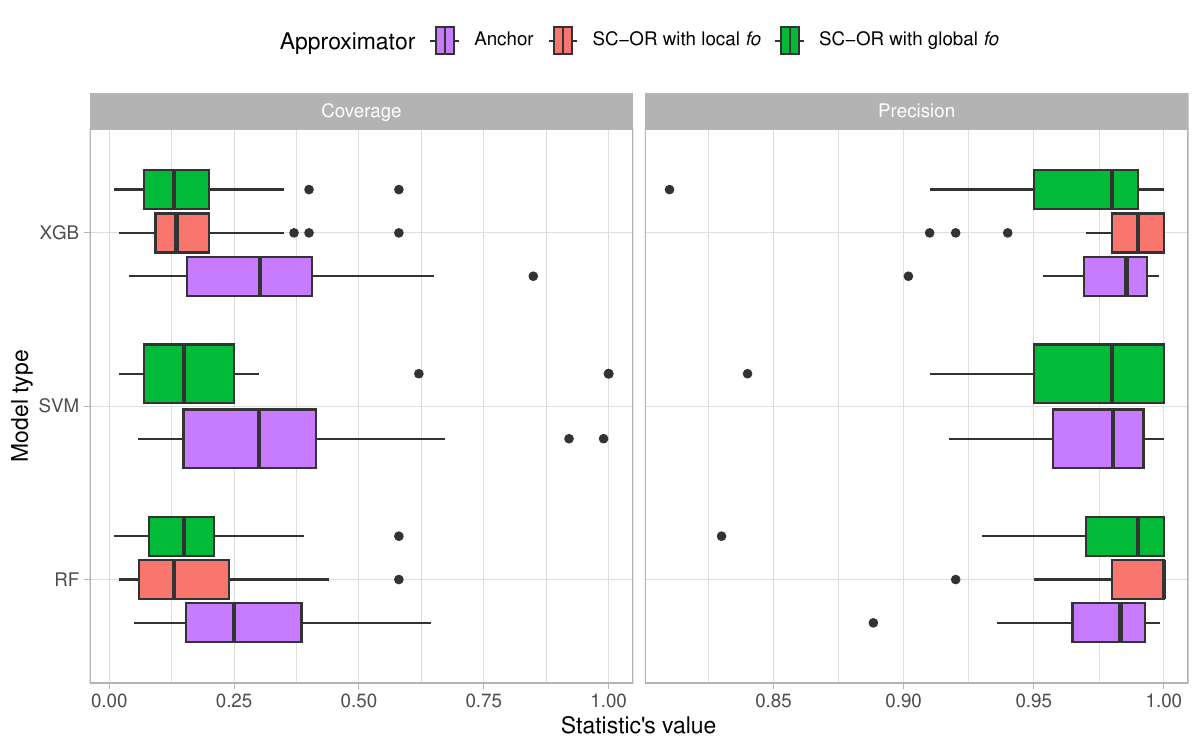}}
    \caption{Quality (precision and coverage) of the rules generated by the Anchors method and approximator that used local and global feature ordering}
    \label{fig:rule_quality_anchors}
\end{figure}

The quality of the rules generated in this experiment is presented in Fig. \ref{fig:rule_quality_anchors}. The results presented in the figure show that rules generated using local importance-based feature ordering (if calculated) have the highest precision. Whereas rules generated by Anchors are characterised by the highest coverage.

\begin{figure}[!htb]
    \centerline{\includegraphics[width=0.8\textwidth]{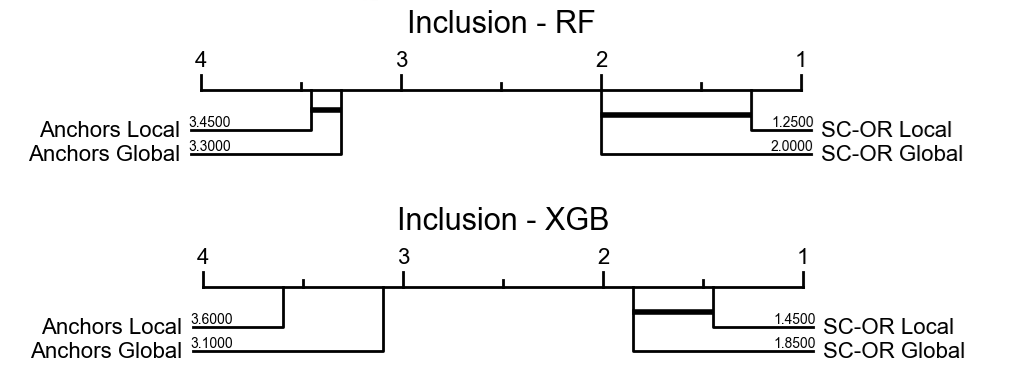}}
    \caption{CD diagrams illustrating the differences between various rule-based explainers in terms of the consistency of mutual feature set inclusion used by the black box model and rule-based explainers.}
    \label{fig:Inclusion}
\end{figure}

\begin{figure}[!htb]
    \centerline{\includegraphics[width=0.8\textwidth]{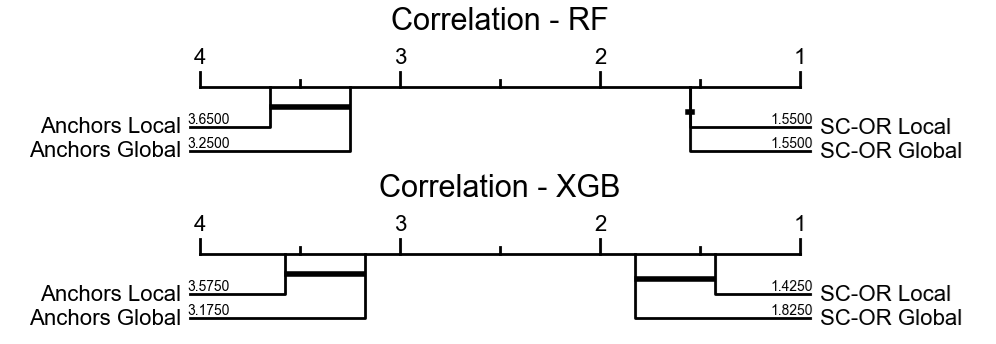}}
    \caption{CD diagrams illustrating the differences between various rule-based explainers in terms of the similarity of feature importance rankings calculated for the black-box model and rule-based explainers.}
    \label{fig:Correlation}
\end{figure}

Figures \ref{fig:Inclusion} and \ref{fig:Correlation} present the CD diagrams illustrating that, for both the Inclusion and Correlation criteria, there are statistically significant differences between the Anchors algorithm and the SC-OR algorithm using importance-based feature ordering generated for the black box model. It was not possible to determine local feature rankings for the SVM models (due to computational time constraints), thus, it was not possible to generate the CD diagrams for this model. A comparison between the Anchors algorithm and the SC-OR method using importance-based feature ordering was performed by means of the Wilcoxon test. In both cases, the method utilizing feature ranking information also demonstrated statistical superiority. The p-values were 0.0006 for Inclusion and 0.001 for Correlation.

\subsection{Case-study}

The case study carried out aims to compare and discus the selected explanations of the XGB model decisions on the selected two datasets. The explanations were generated by the proposed method implemented in the SC-OR algorithm, the Anchors algorithm and the standard rule induction (SC) method.

The first dataset belongs to a medical domain. It was created for the task of identifying genetic aberrations from the results of a flow cytometry technique \cite{jcm11092281}. The presence of aberrations is an important factor for determination of patient prognosis at the initial phase of acute lymphoblastic leukaemia (ALL) diagnostics. ALL is the most frequent leukaemia in children.

The first data instance, which is analysed below, is an example of a true positive decision "no aberrations" of a black box model. The local ordering of features with respect to their importance, which was obtained for this data instance using the SHAP method, has the following form:
cd9, cd123, cd45, cd20, td\_t, cd10, cd22, cd66, cd24, cd13, cd38, c\_ig\_m, cd81, cd34, ng2, cd33, cd15\_65. This ranking consists of 17 attributes, of which cd9 is the most important. For the selected example, rule-based explanations of classifier decision were generated using the SC-OR method with local feature ordering and Anchors. The resulting explanatory rules are presented in Table \ref{tab:cs_all_correct}.

\begin{table}[!htb]
\caption{Explanations of the correct black box model decision on leukemia data - the rules generated by the two methods and the rule characteristics (precision, coverage and average ranking of the attributes creating the rule according to the ranking calculated for the black box model)}
\label{tab:cs_all_correct}
\centering
\begin{tabular}{lp{0.5\linewidth}ccc}
\hline
Method	&	Rule	&	Prec.	&	Cov.	&	Rank.	\\
\hline
SC-OR	&	cd123 = 0 AND cd9 $\in$ [3,10] AND cd10 $\leq$ 34 AND cd22 $\geq$ 2	&	1	&	0.11	&	4	\\
Anchors	&	cd123 = 0 AND cd9 $>$ 1 AND cd34 $\leq$ 4 AND c\_ig\_m $>$ 2 AND cd38 $\leq$ 2	&	1	&	0.06	&	8	\\
SC	&	cd20 $\geq$ 2 AND cd13 $\leq$ 1 &	1	&	0.07	&	7	\\
\hline
\end{tabular}
\end{table}

The rule generated by SC-OR consists of conditions on attributes that are higher in the feature ranking than the attributes used by the Anchors and the standard rule induction (SC) method. In addition, the rule generated by SC-OR is more general, which, for the same precision, gives it more confidence.

The next data instance is an example for which the black box classifier's decision was incorrect - the classifier indicated "aberrations" when it should have decided "no aberrations". The rule-based explanations of the classifier decision, that were generated analogously to the previous example are presented in Table \ref{tab:cs_all_incorrect}. This time, the local ordering of features with respect to their importance, which was obtained for this data instance using the SHAP method, has the following form:
cd22, cd123, cd45, cd34, cd13, td\_t, cd81, c\_ig\_m, cd38, cd9, cd66, cd33, cd10, cd20, ng2, cd24, cd15\_65.

\begin{table}[!htb]
\caption{Explanations of the incorrect black box model decision on leukemia data - the rules generated by the two methods and their characteristics (precision, coverage and average ranking of the attributes used according to the ranking calculated for the black box model)}
\label{tab:cs_all_incorrect}
\centering
\begin{tabular}{p{0.1\linewidth}p{0.5\linewidth}ccc}
\hline
Method	&	Rule	&	Prec.	&	Cov.	&	Rank.	\\
\hline
SC-OR	&	cd123 = 0 AND cd13 $\geq$ 1 AND td\_t $\geq$ 7 AND cd81 $\leq$ 4	&	1	&	0.06	&	5    \\
Anchors	&	cd123 = 0 AND cd9 $>$ 0 AND td\_t $>$ 6 AND cd38 $\leq$ 2 AND cd13 $>$ 0	&	1	&	0.063	&	6.4 \\
SC	&	cd45 $\leq$ 1 AND cd35 $\geq$ 7 AND td\_t $\leq$ 6 AND cd24 $\leq$ 42 AND cd10 $\leq$ 36	&	1	&	0.08	&	9.5 \\
\hline
SC-OR contradictory rule	&   cd22 $\in$ [1,2] AND cd45 $\leq$ 1.00 AND cd34 $\in$ [7,10]   &	0.88	&	0.045	&	2.7 \\
\hline
\end{tabular}
\end{table}

For the explanations presented in Table \ref{tab:cs_all_incorrect}, the rule generated by SC-OR again uses features that are higher in the attribute ranking. Whereas, the properties of the generated rules (precision and coverage) are similar. The contradictory rule reflects the validity of the feature ranking even better than other rules in Table \ref{tab:cs_all_incorrect}. However, its precision is 87\% which, with over 4\% coverage, means that 2 out of 15 examples covered by this rule are from the "aberrations" class. 

The second dataset relates to the customer credit assessment task. It consists of attributes both numerical and categorical. For this dataset, analogous to the analysis presented above, two data instances were analysed.

The first data instance was correctly classified by the black box model as belonging to the "bad" class. 
The local ordering of features with respect to their importance, which was obtained for this data instance using the SHAP method, has the following form:
duration, credit\_amount, checking\_account, purpose, job, housing, age, installment, saving\_accounts, sex.
The resulting explanatory rules generated by the proposed method and the Anchors algorithm are presented in Table \ref{tab:cs_credit_correct}. Again, the rule generated by SC-OR consists of conditions on attributes that are higher in the feature ranking than the attributes used by the Anchors and SC. Furthermore, the rule generated by Anchors consists of a large number of conditions, making it a poorly interpretable explanation of the black box model's decisions.

\begin{table}[!htb]
\caption{Explanations of the correct black box model decision  on credit data - the rules generated by the two methods and the rule characteristics (precision, coverage and average ranking of the attributes creating the rule according to the ranking calculated for the black box model)}
\label{tab:cs_credit_correct}
\centering
\begin{tabular}{lp{0.5\linewidth}ccc}
\hline
Method	&	Rule	&	Prec.	&	Cov.	&	Rank.	\\
\hline
SC-OR	&	duration $\geq$ 44 AND credit\_amount $\in$ [6544, 7118] 	&	1	&	0.02	&	1.5	\\
Anchors	&	duration $>$ 24 AND credit\_amount $>$ 4154.50 AND checking\_account = little AND saving\_accounts = little AND purpose = business AND age $>$ 42 AND job = skilled AND housing = own AND installment $\leq$ 130.74 AND sex = male 	&	1	&	0.008	&	5.7	\\
SC	&	duration $\geq$ 44 AND age $\geq$ 7 AND job = skilled	&	1	&	0.024	&	4.3	\\
\hline
\end{tabular}
\end{table}

The next data instance is an example for which the black box classifier's decision was incorrect - the classifier recommendation was "good", whereas the correct decision was "bad". The local ordering of features with respect to their importance, which was obtained for this data instance using the SHAP method, has the following form: credit\_amount, installment, purpose, housing, job, checking\_account, age, duration, saving\_accounts, sex. The resulting explanatory rules generated by the proposed method and the Anchors algorithm are presented in Table \ref{tab:cs_credit_incorrect}. Additionally, Table \ref{tab:cs_credit_incorrect} contains the contradictory rule generated by SC-OR that explains the decision opposite to that taken by the classifier (this would be the correct decision)

\begin{table}[!htb]
\caption{Explanations of the incorrect black box model decision on credit data - the rules generated by the two methods and the rule characteristics (precision, coverage and average ranking of the attributes creating the rule according to the ranking calculated for the black box model)}
\label{tab:cs_credit_incorrect}
\centering
\begin{tabular}{p{0.1\linewidth}p{0.5\linewidth}ccc}
\hline
Method	&	Rule	&	Prec.	&	Cov.	&	Rank.	\\
\hline
SC-OR	&	installment $\in$ [75.18, 76.12]  	&	1	&	0.114	&	2	\\
Anchors	&	age $>$ 27 AND purpose = radio/TV AND  credit\_amount $\in$ (1380, 4154.50] AND housing = own  	&	0.9	&	0.007	&	3.7	\\
SC	&	purpose = radio/TV AND age $\geq $ 30 AND credit\_amount $\in$ [1341, 1908]  	&	1	&	0.055	&	5	\\
\hline
SC-OR contradictory rule	&	credit\_amount $\in$ [1804, 1843] AND installment $\leq$ 100.69  	&	1	&	0.021	&	1.5	\\
\hline
\end{tabular}
\end{table}

Comparing the rules generated by SC-OR, Anchors and SC, shown in Table \ref{tab:cs_credit_incorrect}, it can be concluded that again, the rule generated by SC-OR consists of conditions on attributes that are higher in the feature ranking than the attributes used by the Anchors and SC methods. In addition, the rule generated by SC-OR is more general and has higher precision. 

Taking the SC-OR contradictory rule into consideration makes explanations difficult to interpret. The explanation provided by SC-OR is unambiguous because it is based on a single attribute: installment $\in$ [75.18, 76.12]. However, the SC-OR contradictory rule also contains this attribute, and the range of the condition built on this attribute (installment $\leq$ 100.69) covers the range [75.18, 76.12]. The SC-OR contradictory rule additionally contains a credit\_amount $\in$ [1804, 1843] condition that refines it. Nevertheless, the SC-OR contradictory rule is very specific and only covers 2\% of the examples in the "bad" class.

\section{Discussion}
\label{sec:discussion}

\subsection{Rule-based explanation schemes}

Rule-based approaches can be used to generate explanations at different levels (global or local), which can be driven by different aspects (model or instance). Therefore, three scenarios were defined and described below, in which rule-based explainability can be applied.

The first scenario concerns global explainability (which can be used for local explanations) for training dataset $D$. 
In this approach, a rule-based approximator $RM$ is generated on the training data on which a black box model $M$ was generated. 
The generated rule-based approximator $RM$ explains the bases for decision-making for $M$ on the training dataset, i.e. it is a source of global explanations. In addition, it allows for local explanations of the training data instances. For each example $x \in D$, an explanation can be obtained based on the rules that cover this example (using only the rules with the decision predicted for this example by the model $M$). The explanation can be based on all rules or only on the best one (with respect to the rule quality measure). In addition, it can be shown whether this data example is covered by other rules with different decisions (contradictory to those that explain example $x$). 
Furthermore, using the RuleXAI package \cite{MACHA2022101209}, it is possible to indicate in the rule set of the $RM$ model the most important conditions that determine the assignment to each decision class.  Similarly, it is possible in the set of rules covering a particular example $x$ to indicate the most important conditions that determined the assignment of this example to that class rather than another. The scenario presented above can be realised for the approximating model $RM$ determined by the covering method (presented in the Algorithm \ref{alg:conquer}) and the object-related method with or without filtering (presented in the Algorithm \ref{alg:conquer_M2}).

\begin{figure}[!htb]
    \centering
    \centerline{\includegraphics[width=0.6\textwidth]{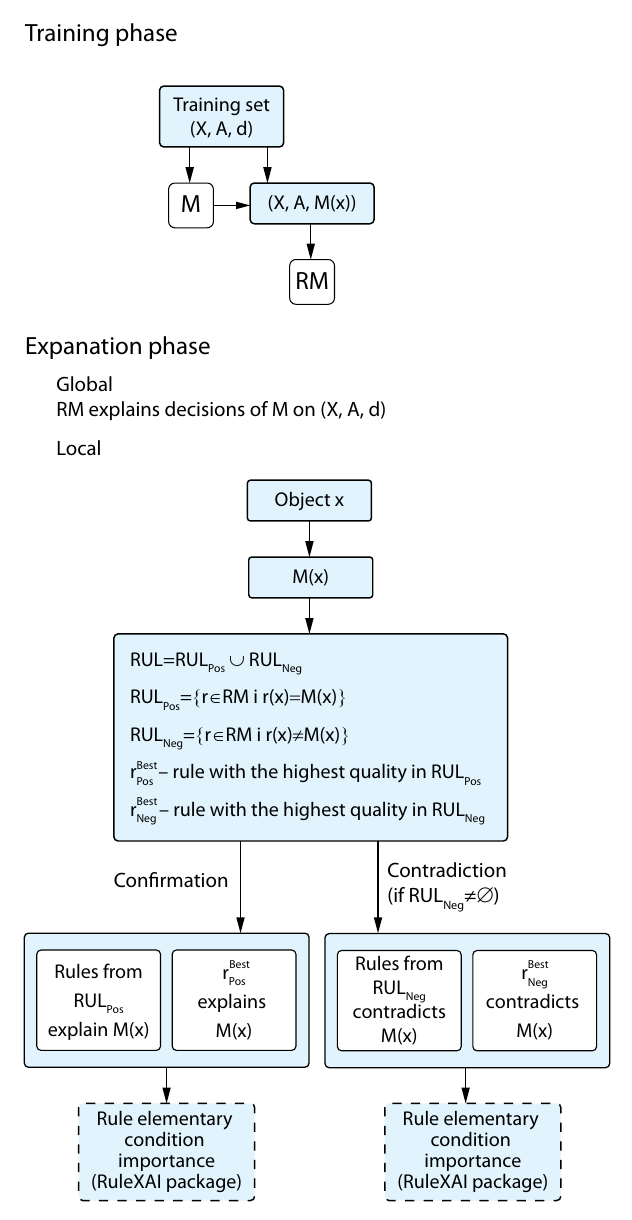}}
    \caption{Block diagram of the rule-based explanation generation scenarios driven by a global model}
    \label{fig:diagram_1}
\end{figure}

The second scenario concerns local explainability driven by a global model. In this case, the explanations being created concern a data example ($x \notin D$) that was not involved in the generation of the black box model $M$ and its rule-based approximator $RM$. Therefore, it concerns the issue of prediction.
To generate the explanation, rules from $RM$ covering $x$ and having the same decision as $M$ for $x$ are selected. The explanation can be based on all rules or only on the best one. Additionally, contradictory rules covering $x$ can be presented. Further, more detailed analysis can be performed by the RuleXAI package as described for the previous scenario.
Both of the above scenarios related to a global rule-based model approximating a black box model are illustrated in Fig. \ref{fig:diagram_1}.

\begin{figure}[!htb]
    \centering
    \centerline{\includegraphics[width=0.5\textwidth]{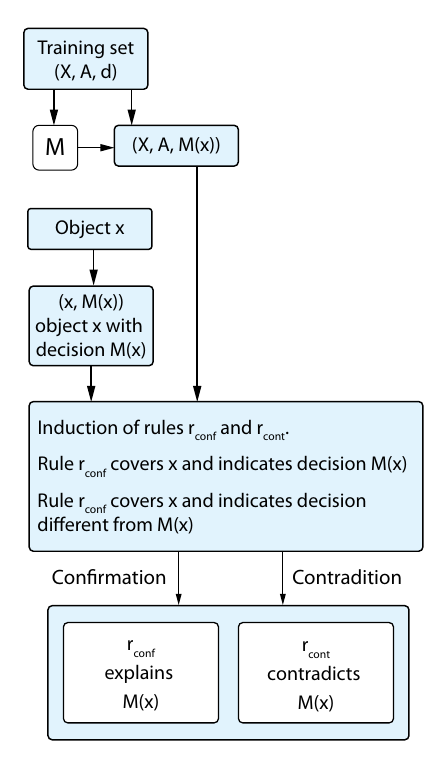}}
    \caption{Block diagram of the rule-based explanation generation scenario driven by a data instance}
    \label{fig:diagram_3}
\end{figure}

Finally, the third scenario concerns local explainability driven by a data instance. Similarly to the previous approach, in this scenario the decision $M(x)$ of the model $M$ for a data instance $x$ that does not belong to the training dataset ($x \notin D$) is explained. This time the approximating model $RM$ is not generated. However, in order to generate an explanation, a rule for $x$ with decision $M(x)$ is generated from the dataset $D$. This rule is generated in such a way that it must cover the data instance $x$ explaining the decision of the model $M$ for this data instance. The explanation of the black box model decision $M(x)$ (generation of a confirmatory rule) can be extended by generating a contradictory rule for the opposite class. In the general case, for a multi-class problem, the generation of a contradictory rule for each class could be considered. The analysis of such explanations could be further supported by information about the quality of the generated rules, which would allow the user to understand the difference in the strength of the explanations presented. A general diagram illustrating the presented scenario is presented in Fig. \ref{fig:diagram_3}. 

For each scenario presented above, it is possible to consider whether and how to take into account the importance-based feature ordering $fo$ introduced in the algorithms presented in the previous section. In each of the scenarios presented,it is possible to disregard the importance-based feature ordering when generating rules and use a basic covering algorithm. Another option is to take into account the global $fo$ ordering, which is identical for each example being explained. Finally, in case of the explainability driven by a data instance, a local ordering can be used, which may be different for each data example. Which approach is used should depend on the user and their subjective purpose and need.

\subsection{Properties of explanations -- rule-based perspective}

To better evaluate the proposed rule-based approach and to present it in a broader context, it has been analysed in terms of the Co-12 properties defined in \cite{10.1145/3583558}. Co-12 allows to assess the quality of explanations from the perspective of their content and presentation and from the perspective of the user. Each of the 12 properties, its key idea given in \cite{10.1145/3583558} and a discussion relating to the properties of rule-based methods in general and the proposed approach in particular are listed below.

Correctness (Nothing but the truth) is addressed by the three conditions (\ref{eq:cond1}-\ref{eq:cond3}) defined in Section \ref{sec:evaluation_conditions}. These conditions define the correctness of the explanations generated by the rule-based approximator of a black box model as the same predictions taken on the basis of the same attributes.
For the conditions defined in Section \ref{sec:evaluation_conditions}, quality measures were selected and used to evaluate the proposed method in the experiments carried out.

Completeness (The whole truth) is implicitly ensured for explanations based on the global model because it is a model generated on the whole training data. 
For explanations based on data instances, it can be ensured by generating an explanatory rule for each data instance. In this case, combining the explanations of all examples will provide a complete representation of the model. Furthermore, the completeness of an explanation based on a single rule can be controlled by setting a threshold for minimum rule precision and coverage.

Consistency (Identical inputs should have identical explanations) property is held by the proposed method, because in its case two data examples with identical feature vectors are referred to the same training dataset. The proposed method does not generate artificial instances in the local neighbourhood of the data example being explained, which would cause differences in explanations. Thus, identical rules will be generated for such two data examples. In addition, however, it is necessary to ensure that the ranking generation method is robust to permutations of examples and attributes.

Continuity (Similar inputs should have similar explanations) property is held by rule-based explanations, which result from the method of rule induction. Rule induction searches for elementary conditions (cuts) in the training dataset, which does not change with the data instances being explained. 
Therefore, if there are no data instances belonging to a different class between the two data examples being explained which would introduce cuts for attribute values, these examples will be covered by one rule. That is, such similar data examples will have the same explanation.
However, in the proposed solution, the explanations additionally depend on the importance-based ranking of the attributes. If this ranking does not change significantly, and small changes in attribute values should not significantly affect their importance, then the generated rules for two similar examples will be identical or similar.

Contrastivity (Answers “why not?” or “what if?” questions) is preserved by rule-based explanations. First of all, two examples classified into different classes will be explained by rules with different conclusions. Furthermore, for different data examples, the rules may differ in the attributes present in the premise part or in the range of elementary conditions where the attributes overlap. It is therefore possible to determine the difference of the rules (this difference is not symmetric), which will indicate which attributes and to what extent are responsible for the distinction of the examples by the model. For the proposed approach, it can be expected that the use of a global feature ranking will make the difference of the determined rules smaller than if a local feature ranking is used. This is because local rankings can be based on more diverse sets of attributes.

Covariate complexity (Human-understandable concepts in the explanation) is assured by explanations in the form of rules, if the attributes analysed represent human-understandable concepts (in this study, by definition, the data has a tabular representation). The rule form of explanations is human-understandable. However, the rule generation algorithm does not determine generalised concepts, although it provides compact explanations by merging overlapping elementary conditions.
Thus, knowing that rules are generated from the features that were used to train the model, it is important that the data representation contains features that will be interpretable to the end user. 

Compactness (Less is more) can be considered for rule explanations from several perspectives. Typically, rules are compact and comprehensible because they have a moderate number of conditions. The compactness of the rules is influenced by the search heuristics used in the pruning phase of rule generation. The more precise the rule should be, the more conditions there will be, while a smaller number of conditions increases the generality of the rule.
Furthermore, in the case of local explanations, for each example to be explained only one rule can be generated as it is performed in the proposed method. In such case the rule covers the example to be explained and has the decision that was returned by the black box model.

Composition (\textit{How} something is explained) property is held by rule-based explanations.
The rule-based presentation format, consisting of a conjunction of conditions leading to a decision presented as a rule conclusion, is considered to be easily interpretable and is therefore used in various XAI methods.

Confidence (Confidence measure of the explanation or model output) is a property that can take into account two components \cite{10.1145/3583558}. A confidence measure of the black box prediction is an issue independent of the rule-based method presented. The second aspect is the determination of confidence for explanations. Precision and coverage measures are commonly used to assess the quality of rules. These measures were determined in the experiments conducted to characterise the quality of the rules (explanations) generated by the proposed method and to compare the quality of the methods used. In addition, it is possible to determine the p-value of a rule or another measure that is based on a contingency table on the basis of these measures. 

Context (How much does the explanation matter in practice?) refers to the extent to which the user and their needs are taken into account to obtain understandable explanations. This aspect was beyond the scope of the presented approach. However, there is an approach presented in \cite{GuideR} that introduces an extension of the rule-based model generation method that is user-driven. In this approach, the user can choose which attributes should be used for rule generation. Using such a method may lead to rules of lower quality. However, it is possible to provide the user with both contextual (user-driven) and automatic explanations, which will give the user a better view of explainability. 

Coherence (Plausibility or reasonableness to users) assesses to what extent the explanation is consistent with relevant background knowledge. 
The coherence defined in this way depends on the performance of the underlying, black box model. If this model is correct, its decision-making process should be based on premises that are coherent with common knowledge. Otherwise, the importance-based ordering of the features generated for the black box model will not be coherent with background knowledge and will affect the form of the rules generated. Moreover, the proposed method depends on the XAI method generating a feature ranking, which may additionally affect the consistency of explanations.
However, the advantage of rule-based methods is that they allow coherence to be verified at different levels of granularity. It is possible to verify the importance of attributes but also the ranges of attribute values \cite{MACHA2022101209}. Furthermore, an advantage of the rule-based representation is its form, in which it is easy for a human expert to capture background knowledge. Representing explanations and knowledge in the same form should facilitate their comparison and verification of coherence.

Controllability (Can the user influence the explanation?) of rule-based explanations can be provided by allowing the user to add their requirements, preferences or domain knowledge. A method that makes this possible using the sequential covering approach to rule generation is presented in \cite{GuideR}. However, allowing the user to influence the explanations generated may result in explanations that do not mimic the operation of the black box model. This would be the opposite result to the motivation for the proposed method.

\section{Conclusions and Future Work}
\label{sec:conclusions}

This paper proposes a method to use external feature ranking to influence rule-based explanations to better mimic the performance of a black box model. The proposed approach is a modification of the sequential covering method, which is used to generate rule-based models. In experiments conducted on a large number of datasets, it was verified that rule-based models are good approximators on both training and test data and can be used as interpretable surrogate models. Furthermore, the analyses carried out allowed to select the implementation of the rule-based model generation method and its main settings.
The rule-based model generated by the proposed approach is compared both with the one generated by the original method, which was created without the importance information provided by the feature ranking of the explained black box model, and with the rules obtained for the Anchors algorithm, which was used as a state-of-the-art method. In addition, the proposed method is illustrated by comparing its results with the Anchors method in a case-study.
The results obtained show that incorporating the feature importance ranking generated for the black box model being explained allows rule-based models to explain black box decisions more reliably (in terms of Inclusion and Correlation).

A detailed report containing the full experimental results, as well as an implementation of the proposed method for which experiments were carried out is publicity available (\href{https://github.com/ruleminer/FI-rules4XAI}{https://github.com/ruleminer/FI-rules4XAI}). 

The proposed method is further discussed to present possible pathways for the application of the rule-based approach in XAI. In addition, it was discussed how rule-based explanations, including the proposed method, relate to Co-12 properties.

The experiments presented were conducted using selected XAI methods generating importance-based feature orderings for the black box model. Depending on the feature importance determination method used, the rules obtained in the proposed method may differ. However, it is assumed that it is up to the user to consciously decide which method they want to use. For the proposed approach, it is also possible to use not a single method but an ensemble of methods generating the feature importance ranking and then combining their results, e.g. by averaging the feature importance rankings and passing such resulting ranking to the rule induction algorithm.

Future work will include extending the functionality of the proposed method to regression and survival analysis tasks, as the induction methods used can operate on such data as well (\cite{gudys2020rulekit}).
In addition, future work will include extending the application of the proposed method to other types of data, in particular image analysis. In this case, the planned approach involves identifying superpixels in the image. Then, based on the groups of superpixels identified in the training images, binary attributes will be defined on which rule induction will be performed.


 \bibliographystyle{elsarticle-num} 
 \bibliography{biblio}

\end{document}